\documentclass[letterpaper]{article}
\usepackage[preprint]{aaai2027}
\usepackage[hyphens]{url}
\usepackage{graphicx}
\usepackage{natbib}
\usepackage{caption}
\usepackage{booktabs}
\usepackage{multirow}
\usepackage{array}
\usepackage{colortbl}
\usepackage{amsmath}
\usepackage{amssymb}

\definecolor{pdmsgray}{gray}{0.90}

\title{Geo-VLA: Geometry-Aware Vision-Language-Action\\
 Planning via Internalization of Map Semantics}
\author{
Ran Chen\equalcontrib,
Jiaxing Ren\equalcontrib,
Zhikun Zhang,
Yunhao Hou,
Junbao Zhuo,
Bochao Zou\corresponding
}
\affiliations{
University of Science and Technology Bei{}jing\\
Bei{}jing, China
}

\begin{document}

\maketitle

\begin{abstract}

Vision-language-action (VLA) models have advanced end-to-end autonomous driving by leveraging foundation models for semantic reasoning and long-tail generalization. However, their planning performance remains limited in complex driving environments because image-only representations inadequately capture planning-relevant road geometry and topology. In this paper, we propose Geo-VLA, a plug-and-play framework that enhances VLA models by learning geometry-aware visual representations. During training, Geo-VLA internalizes geometric map semantics to strengthen road-structure representations, while requiring no HD maps or additional lane information during inference. To support this approach, we introduce Geo-QA, a geometry-focused question-answering dataset that injects road geometry into vision-language representations through contrastive learning and instruction tuning. Experiments on NAVSIM v1 demonstrate that Geo-VLA consistently improves VLA planners with distinct action-generation architectures, achieving 92.1 PDMS and establishing a new state-of-the-art among single-camera VLA planners.

\end{abstract}

\section{Introduction}

Vision-language-action (VLA) models connect vision-language representations with action generation and have emerged as a promising approach to end-to-end autonomous driving \citep{li2025recogdrive,li2025drivevlaw0,jia2026driveworldvla,shang2026dynvla}.
Built on foundation models, VLA models leverage semantic reasoning, visual question answering, instruction following, and long-tail generalization for driving-scene understanding \citep{sima2024drivelm,tian2024drivevlm,shao2024lmdrive}.
Recent methods further use diffusion planning, world-model supervision, or autoregressive action prediction to convert these representations into future ego trajectories \citep{li2025recogdrive,li2025drivevlaw0,jia2026driveworldvla,shang2026dynvla}.
Because the generated trajectories directly affect driving safety, efficiency, and comfort, VLA planning performance depends on whether vision-language representations encode the information required for trajectory generation.

\begin{figure}[t]
  \centering
  \includegraphics[width=1\columnwidth]{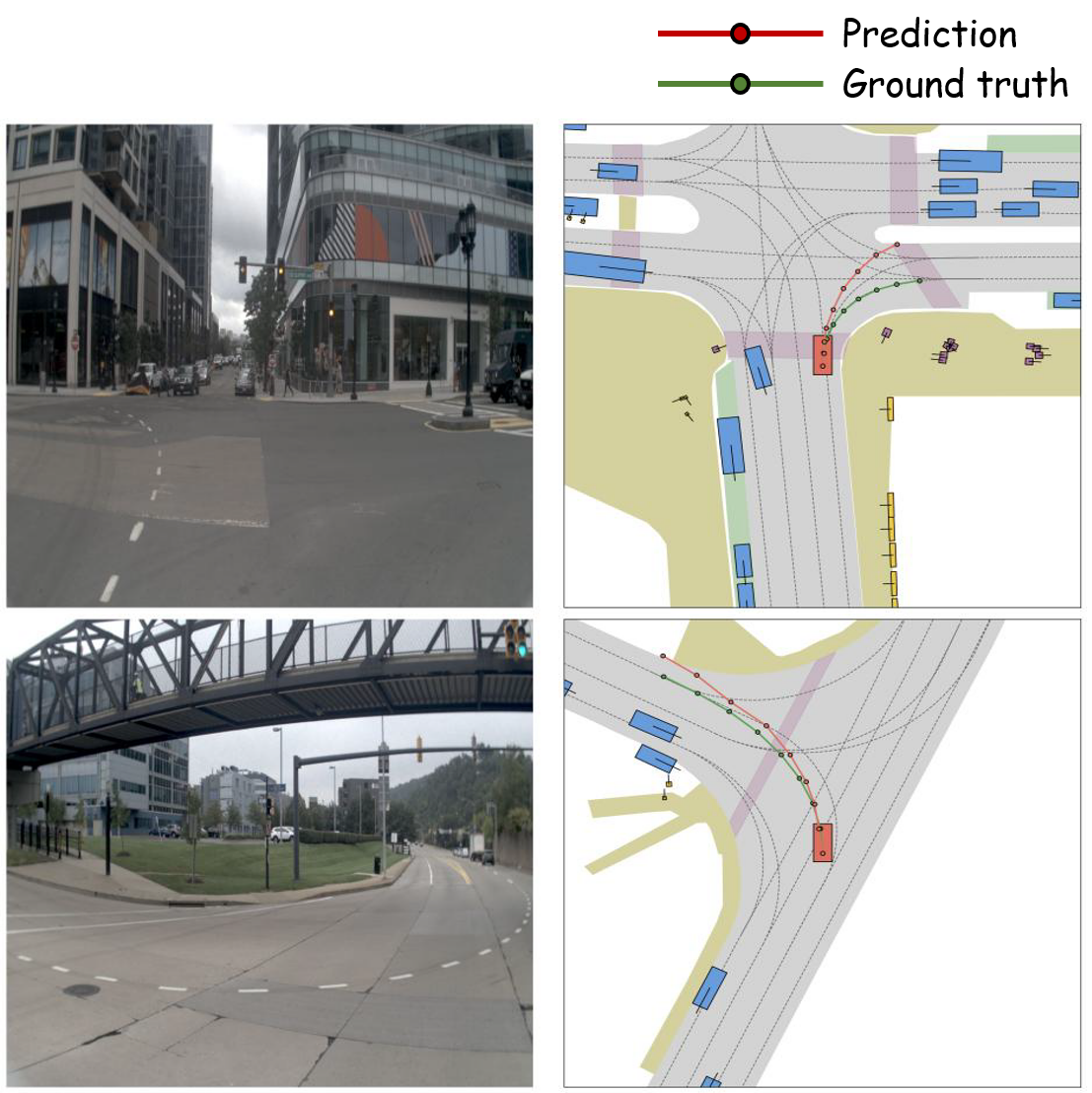}
  \caption{Representative VLA planning failures in turning scenes. Each row
  shows the front-view observation on the left and the corresponding bird's-eye
  view on the right. At intersections and curves, the predicted trajectory
  deviates from the expert trajectory and the intended road geometry.}
  \label{fig:motivation}
\end{figure}

However, existing VLA models obtain road information primarily from camera images, and their visual representations inadequately capture the road geometry and topology needed for trajectory planning, as shown by the representative planning failures in Figure~\ref{fig:motivation}.
Trajectory planning must satisfy static road constraints such as lane boundaries, road curvature, intersection connectivity, drivable-area boundaries, and lane-level topology.
Camera images readily capture the appearance of vehicles and pedestrians, but structured map elements such as lane connectivity and turning restrictions are difficult to infer reliably from road appearance and markings in a single front-view image.
As illustrated in Figure~\ref{fig:motivation}, trajectories predicted by VLA models can deviate from expert trajectories and the intended road geometry in curves and intersections.
These failure cases motivate us to investigate how VLA models can acquire stronger road-structure representations without relying on explicit map inputs during inference.

A direct way to provide such information is to introduce an explicit HD map encoder at inference, which encodes road boundaries, lane connectivity, and drivable areas as structured features for prediction and planning \citep{gao2020vectornet,liao2023maptr}.
However, HD maps require dedicated data collection and annotation, along with ongoing maintenance and updates, because changes in road layouts can render map information outdated \citep{li2021hdmapnet,liu2023vectormapnet,liao2023maptr}.
Their online use also introduces localization, map retrieval, and map encoding into the inference pipeline, increasing system complexity and computational cost.
More importantly, these methods treat map features as additional inputs, leaving the vision-language backbone's limited ability to represent road structure unaddressed while changing the original input setting of the VLA model.
This leads to the central question of this work: can map semantics be used as training supervision so that a VLA model learns road geometry and topology while preserving its original inference interface without map inputs?

To answer this question, we propose Geo-VLA, a plug-and-play framework that enhances VLA models by learning geometry-aware visual representations.
Geo-VLA uses geometric supervision generated from map annotations to adapt the vision-language backbone without redesigning the baseline action decoder.
The adapted backbone is then connected to the original action decoder of each baseline and fine-tuned for trajectory planning under expert-trajectory supervision.
At inference, Geo-VLA preserves the baseline input interface and requires no HD maps, map text, or additional map encoders.

To provide the required supervision, we introduce Geo-QA, a geometry-focused question-answering dataset.
It contains 3,000 samples generated by pairing driving images with descriptions derived from offline map annotations.
Geo-QA covers lane structure, road direction and curvature, intersection structure, drivable-area boundaries, and road topology.
During pretraining, map-derived descriptions form image-text pairs for contrastive learning, while complete question-answer samples are used for instruction tuning.
The two objectives jointly inject planning-relevant road geometry into vision-language representations.
We also construct matched static road-structure and dynamic-object QA variants to examine how different supervision content affects VLA planning.

Our main contributions are summarized as follows.
\begin{itemize}
\item We reveal that existing VLA models inadequately represent static road geometry and topology, and show through ablations with static and dynamic QA supervision that static road-structure supervision provides greater planning benefits than dynamic-object supervision.
\item We propose Geo-VLA, a plug-and-play framework that internalizes planning-relevant map semantics during training while requiring no HD maps, map text, or additional map encoders at inference, and introduce Geo-QA to provide geometry-focused supervision.
\item We conduct experiments on NAVSIM v1 with two open-source VLA baselines and establish a new SOTA among single-camera VLA planners. We further conduct explicit geometric-input comparisons, ablations of static and dynamic QA supervision, and qualitative planning analysis.
\end{itemize}

\begin{figure*}[t!]
  \centering
  \includegraphics[width=0.92\textwidth]{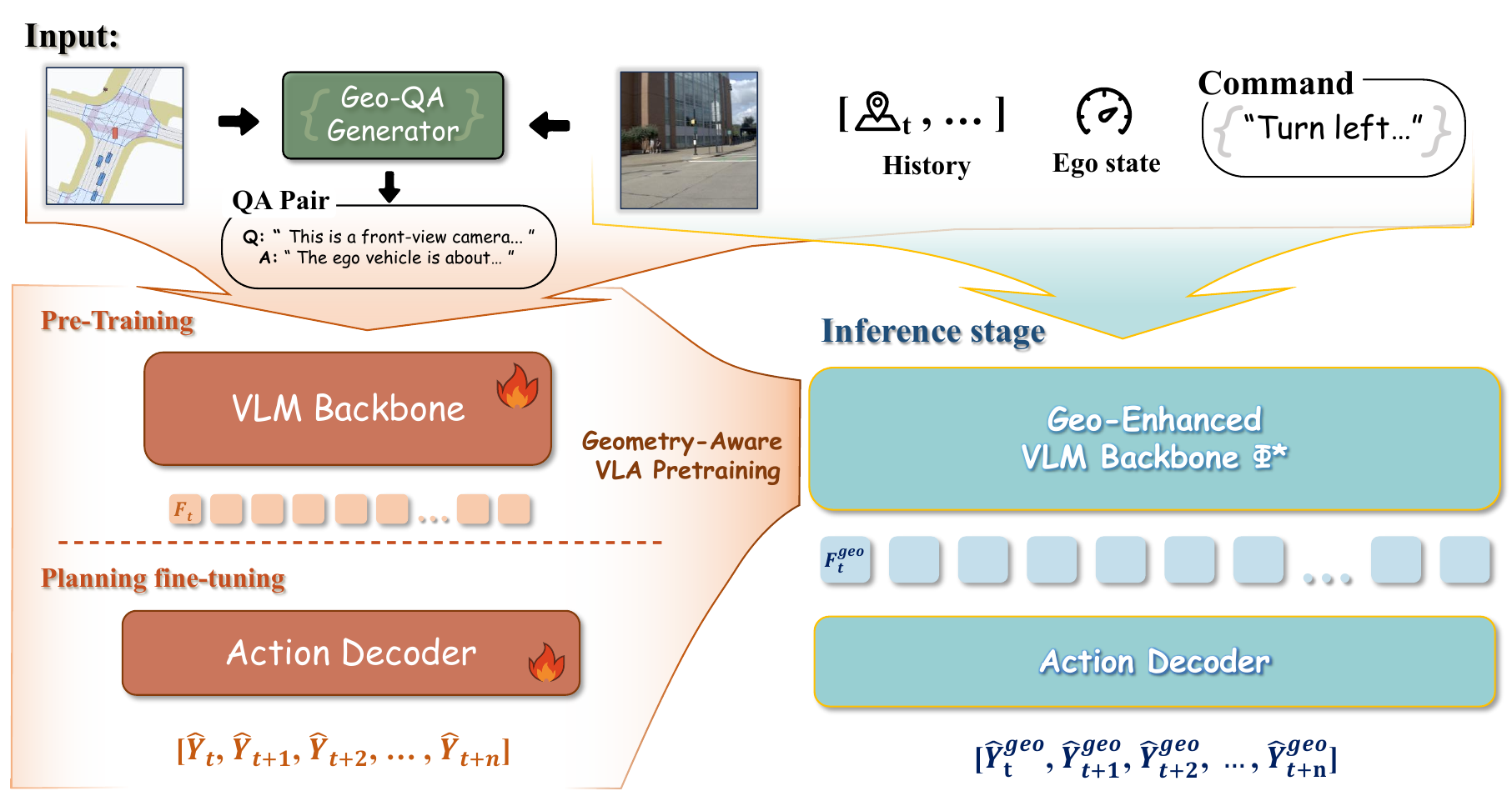}
  \caption{Overview of Geo-VLA for end-to-end VLA planning. During
  geometry-aware pretraining, a front-view image and its paired offline map
  annotation are converted into a Geo-QA pair to adapt the vision-language
  backbone from $\Phi_{\theta}$ to $\Phi^{*}$. The corresponding hidden states
  change from $F_t$ to the geometry-aware representation
  $F_t^{\mathrm{geo}}$. During planning fine-tuning, $\Phi^{*}$ is fixed and
  the original action decoder is optimized with expert-trajectory supervision
  under its original training objective. Here, $\Phi^{*}$ denotes the adapted
  vision-language backbone after pretraining, excluding the auxiliary
  projection heads used only for pretraining. At inference time $t$, the inputs
  remain the front-view image $I_t$, ego history $H_t$, ego state $s_t$, and
  navigation command $c_t$. The decoder predicts
  $\hat{\mathbf{Y}}_t^{\mathrm{geo}}=[\hat{\mathbf{w}}_t,\ldots,
  \hat{\mathbf{w}}_{t+n}]$, where $n$ is the planning horizon. Offline map
  annotations and Geo-QA pairs are used only during training.}
  \label{fig:overview}
\end{figure*}

\section{Related Work}

\subsection{VLA for End-to-End Autonomous Driving Planning}

End-to-end autonomous driving learns to generate future ego trajectories from sensor observations under a planning objective. Existing methods improve this mapping from complementary perspectives. ST-P3, TransFuser, BEVFormer, UniAD, VAD, and VADv2 develop spatiotemporal, multi-view, or planning-oriented scene representations that connect scene understanding with trajectory generation \citep{hu2022stp3,chitta2022transfuser,li2022bevformer,hu2023uniad,jiang2023vad,chen2024vadv2}. Subsequent work further improves temporal modeling, trajectory generation, or policy optimization through long-short-term fusion, multi-target distillation, diffusion generation, reinforcement learning, trajectory scoring, and discrete trajectory modeling. MLSTP, for example, fuses long- and short-term motion cues with a Mamba-based network for trajectory prediction \citep{ren2026mlstp}, while related methods explore the other directions \citep{li2024hydramdp,liao2025diffusiondrive,zou2025diffusiondrivev2,sun2026sparsedrivev2,wang2026reflectdrive2}. Vision-language driving models further use foundation-model knowledge for scene understanding, instruction following, and high-level decision making, while VLA planners connect these representations to action generation through cognitive question answering, predictive or world-model supervision, and dynamics-aware token modeling \citep{sima2024drivelm,tian2024drivevlm,shao2024lmdrive,li2025recogdrive,li2025drivevlaw0,yang2024genad,jia2026driveworldvla,yao2026discretewam,shang2026dynvla}. 
Despite these advances, existing methods mainly strengthen cognition, temporal or dynamics modeling, and action generation, while static road geometry and topology are not dedicated representation-learning targets. 

Geo-VLA addresses this gap before planning fine-tuning and remains compatible with the original action decoder of each baseline.

\subsection{Map and Road-Structure Modeling}

Road geometry and topology provide structural constraints for motion prediction and trajectory planning. Existing methods generally follow two routes to obtain this information. 

The first directly represents map elements or planning-oriented scene structure: VectorNet, UniAD, VAD, and VADv2 model lane geometry, road boundaries, and connectivity as structured features for downstream planning \citep{gao2020vectornet,hu2023uniad,jiang2023vad,chen2024vadv2}. The second reconstructs structured maps online from sensor observations, as in HDMapNet, VectorMapNet, and MapTR \citep{li2021hdmapnet,liu2023vectormapnet,liao2023maptr}. 
The offline route depends on map collection, annotation, maintenance, and updates, whereas the online route adds localization, map construction, retrieval, or encoding to the inference pipeline. Both routes retain explicit map representations or map-construction modules at deployment and therefore change the original input interface of a VLA planner. 

Geo-VLA uses offline map annotations only as training supervision, internalizes the resulting road semantics in the vision-language backbone, and requires no explicit HD map, map text, or additional map encoder at inference.

\subsection{Geometric Supervision and Image-Text Alignment}

Language supervision has been used to provide structured semantics for visual representation learning and driving-scene reasoning. 

DriveLM organizes driving questions around graph-structured scene reasoning, while ReCogDrive uses cognitive question answering to strengthen the connection between visual-language features and planning \citep{sima2024drivelm,li2025recogdrive}. Other VLA studies introduce predictive or world-model supervision and dynamics-aware language or token modeling to improve action reasoning \citep{li2025drivevlaw0,jia2026driveworldvla,yao2026discretewam,shang2026dynvla}. 
These forms of supervision mainly target general scene semantics, cognition, or dynamic behavior; they do not specifically construct training pairs for static road geometry and topology required by trajectory planning. 

Geo-QA focuses language supervision on map-derived road semantics. Its image-text pairs support contrastive learning, and its complete question-answer pairs support instruction tuning, allowing Geo-VLA to internalize planning-relevant geometry during training while preserving the baseline action decoder and inference interface.

\section{Method}

We present Geo-VLA, a plug-and-play training framework that strengthens road-structure representations in VLA planning without redesigning the action decoder. As shown in Figure~\ref{fig:overview}, Geo-VLA first adapts the vision-language backbone through geometry-aware pretraining on Geo-QA, then fine-tunes the original action decoder under the baseline imitation-learning protocol. Section 3.1 defines the task, and Sections 3.2--3.5 describe the framework, Geo-QA construction, pretraining, and planning fine-tuning, respectively.

\begin{figure*}[t!]
  \centering
  \includegraphics[width=0.92\textwidth]{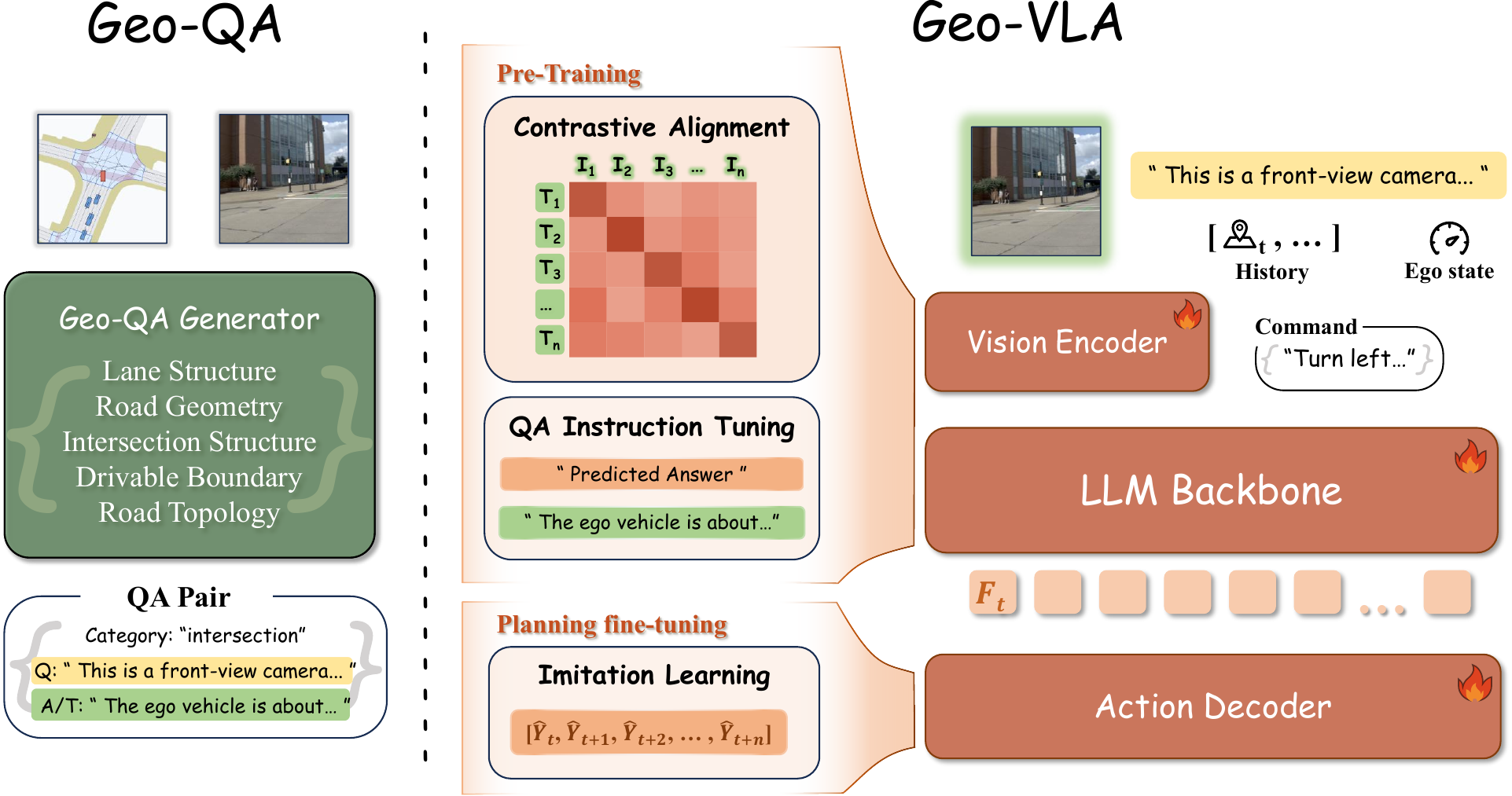}
  \caption{Geo-QA construction and the two-stage training of Geo-VLA. The left
  part shows how the Geo-QA generator uses a front-view image $I_i$, its local
  offline map annotation $\mathcal{M}_i$, and a predefined category $r_i$ to
  obtain a question $Q_i$ and an answer $A_i$ with GPT-5.4. The label A/T
  indicates that the same answer $A_i$ also serves as the textual description
  $T_i$ of map semantics for contrastive learning. The right part presents the
  stage-specific learning objectives and model architecture. Here, $i$
  indexes a Geo-QA sample. Geometry-aware
  pretraining jointly performs contrastive alignment and question-answer
  instruction tuning to optimize the trainable parts of the vision encoder and
  LLM backbone. Planning fine-tuning then optimizes the action decoder through
  the planner's original imitation-learning objective. The adapted backbone produces
  geometry-aware hidden states $F_t^{\mathrm{geo}}$, from which the decoder predicts
  $\hat{\mathbf{Y}}_t^{\mathrm{geo}}$. Flame icons identify the components optimized at each
  stage.}
  \label{fig:pretraining}
\end{figure*}

\subsection{Problem Formulation}

At planning time $t$, the model receives a front-view image $I_t$, an ego history $H_t$, the current ego state $s_t$, and a navigation command $c_t$.
Here, $t$ denotes the current planning time, $I_t$ is the front-view image, $H_t$ is the ego-motion history, $s_t$ is the current ego state, and $c_t$ is the high-level navigation command.
The history contains $m$ ego poses,
\begin{equation}
H_t=[\mathbf{w}_{t-m},\ldots,\mathbf{w}_{t-1}],
\end{equation}
where $j$ indexes a time step, $\mathbf{w}_j=(x_j,y_j,\varphi_j)$ denotes the ego pose at time $j$, $(x_j,y_j)$ is its planar position, and $\varphi_j$ is its heading angle.
The ego state $s_t$ contains motion attributes such as velocity and acceleration, while $c_t$ specifies the high-level navigation command.
The planning target and model prediction contain the waypoint at time $t$ as an anchor followed by $n$ future waypoints,
\begin{equation}
\begin{aligned}
\mathbf{Y}_t&=[\mathbf{w}_t,\mathbf{w}_{t+1},\ldots,\mathbf{w}_{t+n}],\\
\hat{\mathbf{Y}}_t&=[\hat{\mathbf{w}}_t,\hat{\mathbf{w}}_{t+1},\ldots,\hat{\mathbf{w}}_{t+n}],
\end{aligned}
\end{equation}
where $n$ is the number of future prediction steps. Here, $\mathbf{Y}_t$ denotes the expert target trajectory, $\hat{\mathbf{Y}}_t$ denotes the model prediction, and $\hat{\mathbf{w}}_j$ denotes a predicted pose or trajectory waypoint. We use the same pose representation $\mathbf{w}_j$ for historical poses and trajectory waypoints.

We decompose a VLA planner into a vision-language backbone and an action decoder.
The ego context is first converted into a driving prompt,
\begin{equation}
P_t=\mathcal{P}(H_t,s_t,c_t),
\end{equation}
where $\mathcal{P}$ is the prompt-construction function.
The vision-language backbone $\Phi_{\theta}$ encodes the image and prompt into hidden states,
\begin{equation}
F_t=\Phi_{\theta}(I_t,P_t),
\qquad
F_t\in\mathbb{R}^{K_t\times d_h},
\end{equation}
where $F_t$ denotes the sequence of fused vision-language hidden-state tokens produced at planning time $t$ and supplied to the action decoder, $\theta$ denotes all parameters of the vision-language backbone, $\theta_{\mathrm{adpt}}$ denotes the subset of those parameters implemented by trainable adapters, $K_t$ is the number of hidden-state tokens, and $d_h$ is their feature dimension.
The action decoder $\Pi_{\psi}$ maps these hidden states and ego context to a future trajectory,
\begin{equation}
\hat{\mathbf{Y}}_t
=\Pi_{\psi}(F_t,H_t,s_t,c_t),
\end{equation}
where $\psi$ denotes the decoder parameters.
This abstraction does not assume a particular action representation because $\Pi_{\psi}$ includes the planner-specific action generation and final trajectory decoding process.

\subsection{Geo-VLA Framework}

Geo-VLA separates training-time geometric supervision from the inference interface. Offline map annotations adapt the vision-language backbone during training, whereas the baseline action decoder and its deployed inputs are retained at inference. The auxiliary projection heads used for contrastive pretraining are discarded after pretraining and are not part of the deployed backbone $\Phi^{*}$.

As shown in Figure~\ref{fig:pretraining}, Training has two stages. Geometry-aware pretraining optimizes $\theta_{\mathrm{adpt}}$, the parameter-efficient adapters in the vision encoder and LLM backbone, together with the auxiliary projection heads $g_v$ and $g_t$ defined below, yielding the adapted backbone $\Phi^{*}$ while leaving the action decoder fixed \citep{hu2022lora}. The projection heads are used only to define the pretraining objectives and are discarded afterward. Planning fine-tuning then fixes $\Phi^{*}$ and optimizes the original action decoder with expert trajectories under its original protocol.

\subsection{Geo-QA Construction}

Geo-QA converts road relations available in offline maps into geometry-focused supervision for the vision-language backbone. It contains $N=3{,}000$ image-question-answer samples balanced across lane structure, road geometry, intersection structure, drivable-area boundaries, and road topology.

For the $i$-th sample, where $i\in\{1,\ldots,N\}$, we pair a front-view image $I_i$ with its local offline map annotation $\mathcal{M}_i$ and assign a category $r_i$ according to the balanced category allocation.
We then provide $I_i$, $\mathcal{M}_i$, and the category-specific generation instruction to GPT-5.4, which produces a question and a map-grounded answer,
\begin{equation}
(Q_i,A_i)
=\mathcal{G}_{\mathrm{GPT}}(I_i,\mathcal{M}_i,r_i),
\end{equation}
where $\mathcal{G}_{\mathrm{GPT}}$ denotes the constrained question-answer generation process.
The category set is
\begin{equation}
\begin{aligned}
\mathcal{R}&=\{\text{lane},\text{geometry},\text{intersection},
\text{boundary},\text{topology}\},\\
r_i&\in\mathcal{R}.
\end{aligned}
\end{equation}

The category controls data construction and is not provided to the planning model at inference.
For contrastive pretraining, the answer is also used as the textual description of map semantics,
\begin{equation}
T_i=A_i.
\end{equation}
The resulting dataset is
\begin{equation}
\mathcal{D}_{\mathrm{geo}}
=\{(I_i,Q_i,A_i,T_i,r_i)\}_{i=1}^{N}.
\end{equation}
The map annotation $\mathcal{M}_i$ and GPT-5.4 are used only to construct the dataset. Each sample uses the image and question as conversational input and $A_i$ as the target response; together, the five categories describe lateral constraints, curvature, local layouts, and connectivity relevant to feasible trajectories rather than requiring complete map reconstruction.
Detailed category-specific Geo-QA construction, including the corresponding question-answer formats, is provided in Appendix~A.

\begin{table*}[t]
  \centering
  {\small
  \setlength{\tabcolsep}{7.0pt}
  \begin{tabular}{@{}lrrrrr>{\columncolor{pdmsgray}[0pt][0pt]}c@{}}
  \toprule
  Method & NC $\uparrow$ & DAC $\uparrow$ & TTC $\uparrow$ & C $\uparrow$ & EP $\uparrow$ & PDMS $\uparrow$ \\
  \midrule
  ReCogDrive$^\dagger$ {\itshape\citep{li2025recogdrive}} & 98.20 & 97.50 & 94.80 & \textbf{100.00} & 87.50 & 90.8 \\
  DriveFine {\itshape\citep{dang2026drivefine}} & \textbf{98.80} & \textbf{98.60} & \textbf{96.20} & \textbf{100.00} & 86.90 & \underline{91.8} \\
  LaST-VLA {\itshape\citep{luo2026lastvla}} & \underline{98.70} & 97.90 & \underline{95.60} & \textbf{100.00} & \underline{86.80} & 91.3 \\
  DynVLA$^\dagger$ {\itshape\citep{shang2026dynvla}} & 98.00 & 97.20 & 94.20 & \textbf{100.00} & 85.20 & 91.0 \\
  \midrule
  ReCogDrive + Geo-VLA & 98.40 & 98.00 & \underline{95.00} & \textbf{100.00} & \underline{88.00} & 91.2 \\
  DynVLA + Geo-VLA & 98.50 & \underline{98.20} & 95.30 & \textbf{100.00} & \textbf{89.00} & \textbf{92.1} \\
  \bottomrule
  \end{tabular}
  }
  \caption{Comparison with single-camera VLA planners on NAVSIM v1 navtest under the standard single-trajectory protocol. Best-of-$N$ and candidate-selection methods are excluded.
  NC: No at-fault Collisions; DAC: Drivable Area Compliance; TTC:
  Time-to-Collision; C: Comfort; EP: Ego Progress; PDMS: Predictive Driver
  Model Score. $\dagger$ denotes our reproduction. Bold and underline denote the best and second-best results, respectively.}
  \label{tab:main-results}
\end{table*}

\begin{table*}[t]
  \centering
  {\small
  \setlength{\tabcolsep}{3.2pt}
  \begin{tabular}{@{}lllrrrrrr@{}}
  \toprule
  VLA model & Geometry variant & Inference Overhead & NC $\uparrow$ & DAC $\uparrow$ & TTC $\uparrow$ & C $\uparrow$ & EP $\uparrow$ & PDMS $\uparrow$ \\
  \midrule
  \multirow{4}{*}{ReCogDrive}
   & Baseline & Same as baseline & 98.20 & 97.50 & 94.80 & 100.00 & 87.50 & 90.8 \\
   & HD map encoder & High & 98.50 & 99.20 & 95.20 & 100.00 & 88.80 & 91.9 \\
   & Lane detection & Moderate & 98.30 & 98.10 & 94.50 & 100.00 & 88.20 & 91.2 \\
   & \cellcolor{pdmsgray}Geo-QA & \cellcolor{pdmsgray}Same as baseline & \cellcolor{pdmsgray}98.40 & \cellcolor{pdmsgray}98.00 & \cellcolor{pdmsgray}95.00 & \cellcolor{pdmsgray}100.00 & \cellcolor{pdmsgray}88.00 & \multicolumn{1}{>{\columncolor{pdmsgray}[\tabcolsep][0pt]}r@{}}{91.2} \\
  \midrule
  \multirow{4}{*}{DynVLA}
   & Baseline & Same as baseline & 98.00 & 97.20 & 94.20 & 100.00 & 85.20 & 91.0 \\
   & HD map encoder & High & 98.80 & 98.50 & 95.20 & 100.00 & 89.20 & 92.4 \\
   & Lane detection & Moderate & 98.40 & 98.30 & 95.00 & 100.00 & 88.80 & 91.7 \\
   & \cellcolor{pdmsgray}Geo-QA & \cellcolor{pdmsgray}Same as baseline & \cellcolor{pdmsgray}98.50 & \cellcolor{pdmsgray}98.20 & \cellcolor{pdmsgray}95.30 & \cellcolor{pdmsgray}100.00 & \cellcolor{pdmsgray}89.00 & \multicolumn{1}{>{\columncolor{pdmsgray}[\tabcolsep][0pt]}r@{}}{92.1} \\
  \bottomrule
  \end{tabular}
  }
  \caption{Comparison of different map encoding approaches on NAVSIM v1 navtest. Inference Overhead is relative to the baseline: Same as baseline adds no module, Moderate adds online lane detection, and High requires online map localization, retrieval, and encoding. Metric abbreviations follow Table~\ref{tab:main-results}.}
  \label{tab:geometry-ablation}
\end{table*}

\subsection{Geometry-Aware Pretraining with Geo-QA}

Geo-QA provides complementary contrastive and instruction-tuning signals. Question-answer instruction tuning supervises map-grounded responses, while contrastive learning aligns each image with its map-semantic description and separates it from descriptions of other scenes.

For a mini-batch of $B$ image-text pairs $\{(I_b,T_b)\}_{b=1}^{B}$, where $b$ indexes a batch sample, the vision encoder $E_v$ and text path $E_l$ of $\Phi_{\theta}$ produce normalized projection-space features,
\begin{equation}
\begin{aligned}
z_b^{I}&=\operatorname{Norm}\!\left(g_v\!\left(\operatorname{Pool}(E_v(I_b))\right)\right),\\
z_b^{T}&=\operatorname{Norm}\!\left(g_t\!\left(\operatorname{Pool}(E_l(T_b))\right)\right),
\end{aligned}
\end{equation}
where $\operatorname{Pool}(\cdot)$ pools a token sequence, $g_v$ and $g_t$ are auxiliary projection heads, and $\operatorname{Norm}(\cdot)$ denotes $\ell_2$ normalization. We jointly optimize
\begin{equation}
\mathcal{L}_{\mathrm{pre}}
=\frac{1}{2}\left(\mathcal{L}_{\mathrm{i2t}}+\mathcal{L}_{\mathrm{t2i}}\right)
+\lambda\mathcal{L}_{\mathrm{qa}},
\end{equation}
where $\mathcal{L}_{\mathrm{i2t}}$ and $\mathcal{L}_{\mathrm{t2i}}$ are standard in-batch InfoNCE losses for image-to-text and text-to-image retrieval, respectively; $\mathcal{L}_{\mathrm{qa}}$ is the autoregressive answer-prediction loss for $(I_b,Q_b,A_b)$; and $\lambda$ balances the two objectives. The contrastive term organizes road-structure representations across scenes, whereas instruction tuning makes these representations usable for geometry-focused QA. During this stage, $\theta_{\mathrm{adpt}}$, $g_v$, and $g_t$ are optimized while the action decoder remains fixed. Complete loss definitions and pretraining details are provided in the appendix.

\subsection{Planning Fine-Tuning and Inference}

After geometry-aware pretraining, we fix the adapted backbone $\Phi^{*}$ and compute
\begin{equation}
F_t^{\mathrm{geo}}
=\Phi^{*}(I_t,P_t).
\end{equation}
Planning fine-tuning optimizes only the original action decoder, which predicts
\begin{equation}
\hat{\mathbf{Y}}_t^{\mathrm{geo}}
=\Pi_{\psi}
\left(
F_t^{\mathrm{geo}},H_t,s_t,c_t
\right).
\end{equation}
The decoder retains the baseline action representation, planner-specific objective, and optimization protocol; detailed loss formulations and training settings are provided in the appendix. At inference, $\Phi^{*}$ replaces $\Phi_{\theta}$ in the original planner, passing $F_t^{\mathrm{geo}}$ to the fine-tuned decoder through the unchanged baseline interface. Geo-QA construction artifacts, map annotations, map text, and contrastive projection heads are absent, so Geo-VLA requires no HD maps, map text, or additional map encoders at inference.

\section{Experiments}

We evaluate Geo-VLA quantitatively and qualitatively on VLA planners with different action-generation mechanisms.
The experiments further examine whether it preserves the original inference interface, benefits specifically from static road-structure supervision, and produces trajectories that better conform to road geometry.

\begin{figure*}[t]
  \centering
  \includegraphics[width=1\textwidth]{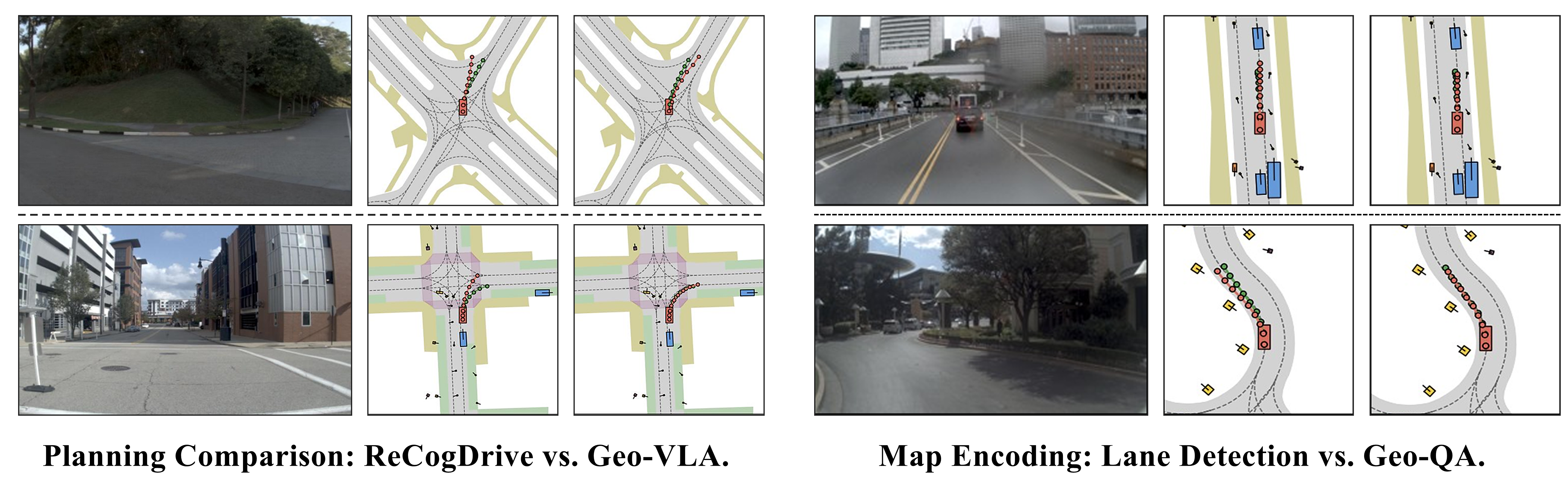}
  \caption{Qualitative planning results in representative turning scenes. In
  the left panel, each row shows the front-view image, ReCogDrive prediction,
  and Geo-VLA prediction from left to right. In the right panel, each row shows
  the front-view image, lane-detection variant, and Geo-QA variant from left to
  right. Red and green denote predicted and expert trajectories, respectively.}
  \label{fig:qualitative}
\end{figure*}

\subsection{Experimental Settings}

\paragraph{Dataset and evaluation protocol.}
We train on NAVSIM v1 navtrain and report results on navtest under the official non-reactive simulation protocol \citep{dauner2024navsim}.
This protocol evaluates predicted ego trajectories against logged observations and annotations without allowing the policy to alter surrounding-agent behavior.

\paragraph{Metrics.}
We report No at-fault Collisions (NC), Drivable Area Compliance (DAC), Time-to-Collision (TTC), Comfort (C), Ego Progress (EP), and the aggregate Predictive Driver Model Score (PDMS).
For evaluation sample $i$, PDMS is computed as
\begin{equation}
\begin{aligned}
\mathrm{PDMS}_i&=\mathrm{NC}_i\,\mathrm{DAC}_i
\frac{5\,\mathrm{EP}_i+5\,\mathrm{TTC}_i+2\,\mathrm{C}_i}{12},\\
\mathrm{PDMS}&=\frac{1}{|\mathcal{S}|}\sum_{i\in\mathcal{S}}\mathrm{PDMS}_i,
\end{aligned}
\end{equation}
where $\mathcal{S}$ denotes the evaluation set.
NC and DAC are multiplicative safety factors, whereas TTC, C, and EP measure temporal safety margin, passenger comfort, and driving progress, respectively.

Following common practice on NAVSIM, PDMS is the principal measure of overall planning performance \citep{dauner2024navsim}. It is computed per scene before averaging and therefore cannot be reproduced by directly combining the averaged component scores in the table.

\paragraph{Implementation details.}
We evaluate Geo-VLA on two open-source VLA baselines, ReCogDrive
\citep{li2025recogdrive}
and DynVLA
\citep{shang2026dynvla}, following the training and evaluation protocols of their original methods.

\subsection{Main Results}

Under the standard single-trajectory NAVSIM v1 protocol, DynVLA + Geo-VLA achieves 92.1 PDMS, establishing a new state of the art among single-camera VLA planners. It exceeds the strongest prior result in Table~\ref{tab:main-results} by 0.3 PDMS. Table~\ref{tab:main-results} compares the two VLA baselines and their Geo-VLA variants with previously reported single-camera VLA results on navtest. Geo-VLA improves ReCogDrive from 90.8 to 91.2 PDMS and DynVLA from 91.0 to 92.1, demonstrating compatibility with distinct action-generation mechanisms. On DynVLA, DAC, TTC, and EP increase from 97.20, 94.20, and 85.20 to 98.20, 95.30, and 89.00, while Comfort remains 100.00. These gains improve road compliance, temporal safety margin, and driving progress without reducing comfort.

\subsection{Ablation Studies}
We conduct two ablation studies that vary geometry integration and QA content while holding the baseline planner, training scale, and evaluation protocol fixed. The first evaluates whether training-time map-semantic supervision can reduce reliance on explicit geometric input at inference, whereas the second evaluates whether the gain is specific to static road structure rather than generic QA content.
\subsubsection{Comparison of Map Encoding Approaches}

To compare the planning benefit and deployment cost of different map encoding approaches, Table~\ref{tab:geometry-ablation} evaluates an HD map encoder, an online lane-detection branch \citep{zheng2022clrnet,zhou2024lane2seq}, and Geo-QA supervision under identical baseline settings. HD map encoding obtains the highest PDMS but incurs high overhead from online localization, retrieval, and encoding. Lane detection removes map retrieval but still introduces moderate overhead through an online detector that processes each input frame. Geo-QA matches lane detection on ReCogDrive and exceeds it by 0.4 PDMS on DynVLA while retaining baseline inference overhead. On DynVLA, it increases TTC and EP from 95.0 and 88.8 to 95.3 and 89.0, preserves maximum Comfort, and changes DAC only marginally from 98.3 to 98.2. Geo-QA remains within 0.7 and 0.3 PDMS of the corresponding HD map variants. Thus, it translates map-semantic supervision into planning performance close to explicit HD map encoding and superior to lane detection without additional inference overhead.

\subsubsection{Static versus Dynamic QA Supervision}

To determine whether Geo-QA benefits planning through static road-structure supervision rather than simply adding an equal amount of driving QA, we pretrain DynVLA with equal-scale dynamic-object QA and static road-structure QA. Dynamic QA describes traffic participants and motion states, whereas static QA describes lanes, road boundaries, intersections, and road topology. Table~\ref{tab:static-dynamic} shows that static QA reaches 92.1 PDMS, exceeding the 91.0 baseline and 90.7 dynamic-QA variant. Because dynamic QA does not improve the baseline under the same data scale, the gain is not explained by sample count or generic QA supervision alone. A plausible explanation is that traffic participants and motion states are more readily captured by the visual semantic priors of VLA backbones, whereas lane connectivity, road boundaries, and intersection structure are less directly learned from trajectory supervision but directly constrain feasible trajectories. Static QA therefore provides complementary supervision for road relations that are weakly represented by the baseline. This comparison does not imply that dynamic agents are unimportant for driving.

\begin{table}[t]
  \centering
  {\small
  \setlength{\tabcolsep}{10pt}
  \begin{tabular}{@{}lcr@{}}
  \toprule
  Model & QA type & PDMS $\uparrow$ \\
  \midrule
  \multirow{3}{*}{DynVLA} & None & 91.0 \\
   & Dynamic & 90.7 \\
  & \cellcolor{pdmsgray}Static & \multicolumn{1}{>{\columncolor{pdmsgray}[\tabcolsep][0pt]}r@{}}{92.1} \\
  \bottomrule
  \end{tabular}
  }
  \caption{Effect of static and dynamic QA supervision.}
  \label{tab:static-dynamic}
\end{table}

\subsection{Qualitative Analysis}

Figure~\ref{fig:qualitative} illustrates representative turning and curved-road scenes. In Figure~\ref{fig:qualitative}(a), ReCogDrive cuts toward the inside of the turn and departs from the expert route, whereas Geo-VLA follows the change in heading and remains closer to the intended lane corridor. In Figure~\ref{fig:qualitative}(b), Geo-QA produces trajectories closer to the expert route than the lane-detection variant, most visibly in the curved-road scene where it follows the road curvature without drifting inward. This advantage is consistent with how the two schemes supply geometry: online lane detection adds local lane cues and a separate inference branch, whereas Geo-QA uses map-grounded QA to supervise lane structure, curvature, and connectivity during training. The resulting backbone can encode road constraints more effectively for the decoder, yielding road-aligned trajectories without additional inference overhead.

\section{Conclusion}

We address limited representations of planning-relevant road geometry and topology in VLA planning. We propose Geo-VLA, a plug-and-play framework, together with Geo-QA, a geometry-focused supervision dataset. Geo-VLA combines image-text contrastive learning and QA instruction tuning to internalize map semantics during training, while retaining the original action decoder and requiring no HD maps, map text, or additional map encoders at inference. Under the standard single-trajectory NAVSIM v1 protocol, Geo-VLA consistently improves ReCogDrive and DynVLA, achieving 92.1 PDMS, establishing a new state of the art among single-camera VLA planners. The map encoding comparison further shows that Geo-VLA approaches HD map performance and surpasses lane detection on DynVLA while retaining baseline inference cost. One potential limitation is that Geo-VLA relies on offline map annotations to construct Geo-QA, so their coverage and quality bound the training-time supervision. Furthermore, our evaluation is limited to single-camera planners under non-reactive protocol of NAVSIM, behavior under interactive traffic remains to be established.

\FloatBarrier

\bibliography{references}

\clearpage
\appendix

\section{Geo-QA Construction and Sampling}
\label{app:geoqa-construction}

We select NAVSIM v1 because it is a widely adopted benchmark for end-to-end
driving planning whose standardized non-reactive protocol and unified planning
metrics enable direct comparison with recent VLA planners. Geo-QA is constructed
from sampled NAVSIM front-view frames and their local offline map annotations.
The sampling process is category-balanced: the final
dataset contains 3,000 samples, with 600 samples for each of the five
road-geometry and topology categories. For each sampled frame, we associate the
front-view image $I_i$ with a local map annotation $\mathcal{M}_i$ and a
category label $r_i$. The category label is used only to select the generation
instruction and balance the dataset; it is not used by the VLA planner at
inference.

Given $I_i$, $\mathcal{M}_i$, and $r_i$, GPT-5.4 generates a question $Q_i$ and
a map-grounded answer $A_i$ under a category-specific instruction. The
instruction requires the answer to describe planning-relevant road relations
supported by the map annotation, rather than general image captions or object
lists. The answer $A_i$ is used in two ways during pretraining: it is the target
response for question-answer instruction tuning, and it is also treated as the
map-semantic text $T_i$ for image-text contrastive learning.

\begin{center}
\scriptsize
\setlength{\tabcolsep}{4pt}
\begin{tabular}{p{0.22\columnwidth}p{0.67\columnwidth}}
\toprule
Category & Map semantics and representative question form \\
\midrule
Lane structure & Lane count, ego-lane index, lane-line type, and lateral constraints. \emph{Question form:} How many lanes are present, and which lane contains the ego vehicle? \\
Road geometry & Straight or turning road shape, curve direction, and curvature. \emph{Question form:} Is the road straight or turning, and in which direction does it curve? \\
Intersection structure & Intersection type, distance, and local road layout. \emph{Question form:} What intersection structure lies ahead, and how far away is it? \\
Drivable-area boundaries & Road boundary, lane width, curb, shoulder, and off-road risk. \emph{Question form:} How far is the ego vehicle from the left and right road boundaries? \\
Road topology & Lane merge, split, allowed maneuver, and downstream connectivity. \emph{Question form:} Does the current lane merge or split, and which maneuvers are allowed? \\
\bottomrule
\end{tabular}
\captionof{table}{Geo-QA category design. The categories are used for balanced data
construction and cover road relations that constrain feasible trajectories.}
\label{tab:app-geoqa-categories}
\end{center}

Table~\ref{tab:app-geoqa-categories} summarizes the five categories. They are
chosen to cover complementary map semantics: lane and boundary questions
describe lateral constraints, road-geometry questions describe direction and
curvature, intersection questions describe local layouts, and topology
questions describe connectivity, merges, and splits. Geo-QA does not ask the
model to reconstruct a complete HD map. Instead, each sample extracts a local
road relation that affects the feasible region or direction of the planned
trajectory.

\FloatBarrier

\section{Geo-QA Data Format}
\label{app:geoqa-format}

Geo-QA is stored as a conversation-style JSONL file in which each line
represents one training sample. The file used in our experiments contains
3,000 records, and the \texttt{image} field points to a sampled NAVSIM
front-view frame under \texttt{map\_qa\_images/}. Each record contains sample
metadata and a two-turn conversation. Table~\ref{tab:app-jsonl-schema} maps the
main fields to their roles in training using a representative record.
The underlying conversation keeps the standard \texttt{from}/\texttt{value}
structure, with the user turn marked as \texttt{human} and the target turn
marked as \texttt{gpt}. The serialized category identifiers follow the category
set defined in the main text: \texttt{lane}, \texttt{geometry},
\texttt{intersection}, \texttt{boundary}, and \texttt{topology}.

\begin{center}
\scriptsize
\setlength{\tabcolsep}{4pt}
\begin{tabular}{p{0.24\columnwidth}p{0.65\columnwidth}}
\toprule
Field & Meaning and example \\
\midrule
\texttt{id} & Unique sample identifier, e.g., \texttt{mapqa\_001910}. \\
\texttt{image} & Relative path to the sampled front-view frame $I_i$, stored as \texttt{map\_qa\_images/<token>.jpg}. \\
\texttt{category} & Category $r_i$ for balanced construction and analysis, e.g., \texttt{boundary}. \\
Human turn & \texttt{from=human}; \texttt{value} contains \texttt{<image>} and the question $Q_i$. \\
Assistant turn & \texttt{from=gpt}; \texttt{value} contains the map-grounded answer $A_i$ and supplies the text $T_i$. \\
\bottomrule
\end{tabular}
\captionof{table}{Serialization of one Geo-QA record. The schema separates metadata
from the two-turn conversation and makes explicit how each field is used during
pretraining. The category field is not provided to the deployed planner.}
\label{tab:app-jsonl-schema}
\end{center}

\begin{figure}[t]
\centering
\begin{minipage}{\linewidth}
\centering
\includegraphics[width=\linewidth]{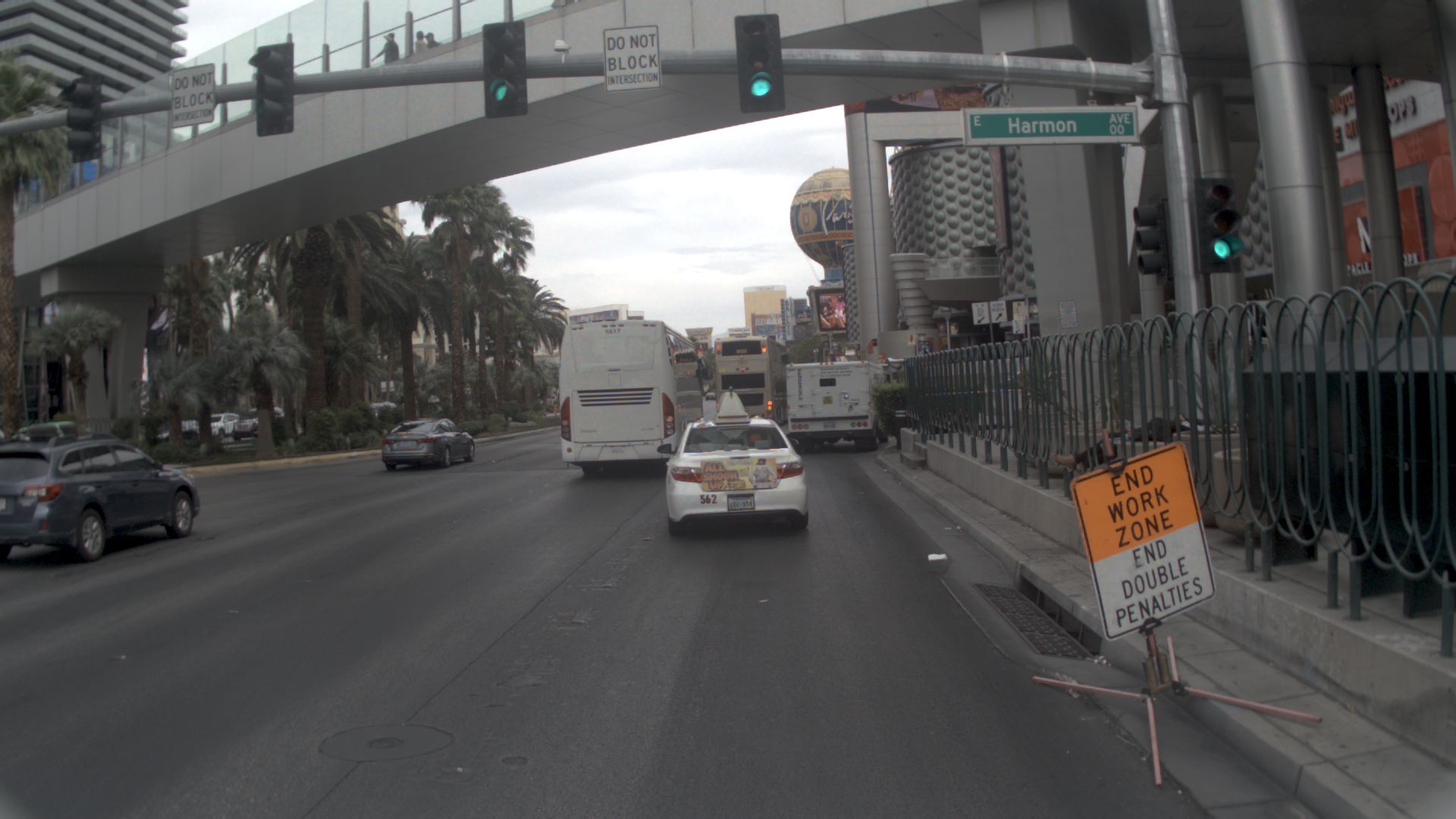}
\raggedright\small
\textbf{ID / Category:} \texttt{mapqa\_001910} / \texttt{boundary}\\
\textbf{Question:} Approximately how far is the ego vehicle from the left and
right road boundaries?\\
\textbf{Answer:} The ego vehicle is about 2.9 m from the left road boundary and
about 2.5 m from the right road boundary.
\end{minipage}
\caption{A representative Geo-QA record shown in reader-facing form. The sample
pairs a sampled front-view image with a geometry-focused question and a
map-grounded answer.}
\label{fig:app-geoqa-examples}
\end{figure}

Figure~\ref{fig:app-geoqa-examples} presents one record in a reader-facing form.
This view exposes the complete training unit without requiring the reader to
parse JSON syntax: the image and question form the conversational input, while
the map-grounded answer is the prediction target. The same answer is reused as
map-semantic text for contrastive learning.

\FloatBarrier

\section{Static and Dynamic QA Probe}
\label{app:qa-probe}

The planning ablation in the main text shows that static road-structure QA is
more beneficial than dynamic-object QA on DynVLA. As a complementary analysis,
we conduct a QA probe on ReCogDrive to examine whether this difference is
associated with the model's understanding of static road structure rather than
with the addition of QA supervision alone. Under the same data scale, we compare
the baseline with variants trained using dynamic QA and static QA. The static
probe covers lane structure, road boundaries,
intersections, curvature, and topology, whereas the dynamic probe covers traffic
participants and motion states. We report GPT-Score on a 0--100 scale, where a
higher value indicates better agreement with the reference answer.

\begin{center}
\small
\setlength{\tabcolsep}{5.5pt}
\begin{tabular}{lrrr}
\toprule
Method & Static & Dynamic & Avg. \\
\midrule
ReCogDrive & 63.10 & 70.30 & 66.70 \\
Dynamic QA & 61.10 & 72.90 & 67.00 \\
Static QA & 75.10 & 73.60 & 74.35 \\
\bottomrule
\end{tabular}
\captionof{table}{QA probe results measured by GPT-Score. Dynamic QA and static
QA use the same number of training samples.}
\label{tab:app-probe}
\end{center}

As shown in Table~\ref{tab:app-probe}, dynamic QA increases the dynamic score
from 70.30 to 72.90 but reduces the static score from 63.10 to 61.10, yielding
only a marginal average improvement from 66.70 to 67.00. In contrast, static QA
raises the static score to 75.10 and the dynamic score to 73.60, increasing the
average score to 74.35. A possible explanation is that vehicles, pedestrians,
and their motion states are directly observable and already well represented by
the VLA backbone, leaving limited room for dynamic QA to improve the missing road
constraints. Static QA instead provides explicit supervision for lane relations,
road boundaries, and topology that are difficult to infer from visual appearance
alone. The probe therefore supports the planning ablation: the gain of Geo-QA is
primarily associated with improved static road-structure understanding rather
than generic QA supervision.

\section{Complete Objective Definitions}
\label{app:objective-details}

As a supplement to the methodology, this section expands the loss definitions
used in Sections~3.4 and 3.5. For a mini-batch of $B$ Geo-QA samples, the image-text pairs are
\mbox{$\{(I_i,T_i)\}_{i=1}^{B}$}. The vision encoder $E_v$ and text path $E_l$
produce pooled features
\begin{equation}
h_i^{I}=\operatorname{Pool}(E_v(I_i)),\qquad
h_i^{T}=\operatorname{Pool}(E_l(T_i)),
\end{equation}
where $\operatorname{Pool}(\cdot)$ converts a token sequence into one feature
vector. The auxiliary projection heads $g_v$ and $g_t$ map the two features
into a shared $d_z$-dimensional embedding space:
\begin{equation}
z_i^{I}=\operatorname{Norm}(g_v(h_i^{I})),\qquad
z_i^{T}=\operatorname{Norm}(g_t(h_i^{T})).
\end{equation}
Here $\operatorname{Norm}(\cdot)$ denotes $\ell_2$ normalization, and
$z_i^{I},z_i^{T}\in\mathbb{R}^{d_z}$. The scaled similarity between image $i$
and text $j$ is
\begin{equation}
s_{ij}=\frac{(z_i^{I})^\top z_j^{T}}{\tau},
\end{equation}
where $\tau>0$ is the temperature. The matched image-text pair shares the same
sample index, so $s_{ii}$ is the positive similarity and $s_{ij}$ with
$j\ne i$ forms in-batch negative similarities.

The image-to-text and text-to-image InfoNCE losses are
\begin{equation}
\mathcal{L}_{\mathrm{i2t}}
=-\frac{1}{B}\sum_{i=1}^{B}
\log
\frac{\exp(s_{ii})}{\sum_{j=1}^{B}\exp(s_{ij})},
\end{equation}
\begin{equation}
\mathcal{L}_{\mathrm{t2i}}
=-\frac{1}{B}\sum_{i=1}^{B}
\log
\frac{\exp(s_{ii})}{\sum_{j=1}^{B}\exp(s_{ji})}.
\end{equation}
Their symmetric contrastive objective is
\begin{equation}
\mathcal{L}_{\mathrm{con}}
=\frac{1}{2}
\left(
\mathcal{L}_{\mathrm{i2t}}
+\mathcal{L}_{\mathrm{t2i}}
\right).
\end{equation}

For the QA instruction-tuning objective, the answer
$A_i=(a_{i,1},\ldots,a_{i,L_i})$ contains $L_i$ answer tokens. The
autoregressive loss is
\begin{equation}
\mathcal{L}_{\mathrm{qa}}
=-\frac{1}{\sum_{i=1}^{B}L_i}
\sum_{i=1}^{B}\sum_{\ell=1}^{L_i}
\log p_{\theta}
\left(a_{i,\ell}\mid I_i,Q_i,a_{i,<\ell}\right),
\end{equation}
where $a_{i,<\ell}$ denotes the answer prefix before token $\ell$, and
$p_{\theta}$ is the conditional token distribution of the vision-language
backbone. The complete pretraining objective is
\begin{equation}
\mathcal{L}_{\mathrm{pre}}
=\mathcal{L}_{\mathrm{con}}
+\lambda\mathcal{L}_{\mathrm{qa}},
\end{equation}
where $\lambda$ balances contrastive alignment and question-answer instruction
tuning.

After pretraining, the adapted backbone $\Phi^{*}$ produces
\begin{equation}
F_t^{\mathrm{geo}}=\Phi^{*}(I_t,P_t).
\end{equation}
Let $\mathcal{D}_{\mathrm{plan}}$ denote the planning dataset associated with
the selected VLA planner, and let
\mbox{$\xi_t=(I_t,H_t,s_t,c_t,\mathbf{Y}_t)$} denote one planning sample.
Define the planner input tuple
\mbox{$\mathbf{x}_t^{\mathrm{plan}}=(F_t^{\mathrm{geo}},H_t,s_t,c_t)$}.
Geo-VLA preserves the planner-specific action representation and
imitation-learning objective, which we write in the generic form
\begin{equation}
\mathcal{L}_{\mathrm{plan}}(\psi)
=\mathbb{E}_{\xi_t\sim\mathcal{D}_{\mathrm{plan}}}
\left[
\ell_{\mathrm{base}}\big(
\Pi_{\psi}(\mathbf{x}_t^{\mathrm{plan}}),
\mathbf{Y}_t
\big)
\right],
\end{equation}
\begin{equation}
\psi^{*}=\arg\min_{\psi}\mathcal{L}_{\mathrm{plan}}(\psi).
\end{equation}
Here $\ell_{\mathrm{base}}$ is the original planning loss defined by the
selected planner, and $\psi^{*}$ is the fine-tuned decoder parameter set.
Geo-VLA does not impose an additional planner loss beyond the selected
baseline's original objective.

Following the two-stage training protocol in the main text, geometry-aware
pretraining uses rank-16 LoRA adapters as $\theta_{\mathrm{adpt}}$ in the vision
encoder and LLM backbone. This stage optimizes $\theta_{\mathrm{adpt}}$, $g_v$,
and $g_t$ while keeping the action decoder fixed. The projection heads are then
discarded, and the adapted backbone $\Phi^{*}$ is fixed during planning
fine-tuning. Only the decoder parameters $\psi$ are optimized in the second
stage, with input preprocessing, action representation, planning loss,
optimization schedule, and evaluation protocol following the selected
baseline.

\end{document}